# Robust Metric Learning by Smooth Optimization

**Kaizhu Huang**
NLPR,
Institute of Automation
Chinese Academy of Sciences
Beijing, 100190 China

**Rong Jin**
Dept. of CSE
Michigan State University
East Lansing, MI 48824

**Zenglin Xu**
Saarland University
& MPI for Informatics
Campus E14, 66123
Saarbrucken, Germany

**Cheng-Lin Liu**
NLPR
Institute of Automation
Chinese Academy of Sciences
Beijing, 100190 China

## Abstract

Most existing distance metric learning methods assume perfect side information that is usually given in pairwise or triplet constraints. Instead, in many real-world applications, the constraints are derived from side information, such as users' implicit feedbacks and citations among articles. As a result, these constraints are usually noisy and contain many mistakes. In this work, we aim to learn a distance metric from noisy constraints by robust optimization in a worst-case scenario, to which we refer as **robust metric learning**. We formulate the learning task initially as a combinatorial optimization problem, and show that it can be elegantly transformed to a convex programming problem. We present an efficient learning algorithm based on smooth optimization [7]. It has a worst-case convergence rate of $\mathcal{O}(1/\sqrt{\varepsilon})$ for smooth optimization problems, where $\varepsilon$ is the desired error of the approximate solution. Finally, our empirical study with UCI data sets demonstrate the effectiveness of the proposed method in comparison to state-of-the-art methods.

## 1 Introduction

Distance metric learning is an important fundamental research topic in machine learning. A number of studies [1, 9, 8, 15, 12, 13, 6] have shown that with appropriately learned distance metrics, we will be able to improve significantly both the classification accuracy and the clustering performance. Depending on the nature of the side information, we can learn a distance metric either from pairwise constraints or from triplet constraints. In this work, we focus on distance metric learning from triplet constraints because of its empirical success. It aims to learn a distance function $f$ with a metric matrix $A$ for a data set that is consistent with a set of triplet constraints $\mathcal{D} = \{(\mathbf{x}_i, \mathbf{y}_i, \mathbf{z}_i) | f(\mathbf{x}_i, \mathbf{y}_i) \leq f(\mathbf{x}_i, \mathbf{z}_i), i = 1, \ldots N\}$ [6]. In the simplest form, the distance function $f$ associated with $A$ between a pair of data points $(\mathbf{x}_i, \mathbf{x}_j)$ is computed as $(\mathbf{x}_i - \mathbf{x}_j)^\top A(\mathbf{x}_i - \mathbf{x}_j)$. For each triplet $(\mathbf{x}_i, \mathbf{y}_i, \mathbf{z}_i)$ in $\mathcal{D}$, we assume instance $\mathbf{x}_i$ is more similar to $\mathbf{y}_i$ than to $\mathbf{z}_i$, leading to a smaller distance between $\mathbf{x}_i$ and $\mathbf{y}_i$ than that between $\mathbf{x}_i$ and $\mathbf{z}_i$.

Most metric learning algorithms assume perfect side information (i.e., perfect triplet constraints in our case). This however is not always the case in practice because in many real-world applications, the constraints are derived from the side information such as users' implicit feedbacks and citations among articles. As a result, these constraints are usually noisy and consist of many mistakes. We refer to the problem of learning distance metric from noisy side information as **robust distance metric learning**. Feeding the noisy constraints directly into a metric learning algorithm will inevitably degrade its performance, and more seriously, it may even result in worse performance than the straightforward Euclidean distance metric, as demonstrated in our empirical study.

To address this problem, we propose a general framework of robust distance metric learning from noisy or uncertain side information in this paper. We formulate the learning task initially as a combinatorial integer optimization problem, each $\{0,1\}$ integer variable indicating whether the triplet constraint is a noisy or not. We then prove that this problem can be transformed to a semi-infinite programming problem [4] and further to a convex op-timization problem. We show that the final problem can be solved by Nesterov's smooth optimization method [7], which has a worst-case convergence rate of $\mathcal{O}(1/\sqrt{\varepsilon})$ for smooth objective functions, where $\varepsilon$ is required level of errors. This is significantly faster than the sub-gradient method whose convergence rate is usually $O(1/\varepsilon^2)$. The proposed method may also be adapted to the other distance metric learning algorithms, e.g., the Large Margin Nearest Neighbor (LMNN) method [12], when the side information is noisy.

The rest of this paper is organized as follows. In the next section, we will briefly discuss the related work. In Section 3, we review distance metric learning with perfect side information. We then introduce the robust metric learning framework in Section 4. In Section 5, we evaluate our proposed framework on five real world data sets. Finally, we set out the conclusion in Section 6 with some remarks.

## 2 Related Work

Many algorithms have been proposed for distance metric learning in the literature. Exemplar algorithms include Information-Theoretic Metric Learning [3] (ITML), Relevant Component Analysis (RCA) [1], Dissimilarity-ranking Vector Machine (D-ranking VM) [8], Generalized Sparse Metric Learning (GSML) [5], the method proposed by Xing et al. [15], and Large Margin Nearest Neighbor (LMNN) [12, 13]. In particular, the last two methods are regarded as state-of-the-art approaches in this field. Specifically, ITML maps the metric space to a Gaussian distribution space and searches the best metric by learning the optimal Gaussian that is consistent with the given constraints. [11] extends this work by the idea of information geometry. In [15], a distance metric is then learned to shrink the averaged distance within must-links while enlarging the average distance within cannot-links simultaneously. In [1], RCA seeks a distance metric which minimizes the covariance matrix imposed by the equivalence constraints. In [12, 13], LMNN is proposed to learn a large margin nearest neighbor classifier. Huang et al. proposed a unified sparse distance metric learning method [5] that provides an interesting viewpoint for understanding many sparse metric learning methods. Besides the empirical results, the generalized performance of regularized distance metric learning was analyzed in the recent study [6]. More discussion of distance metric learning can be found in the survey [17].

All the above methods assume perfect side information. In case of noisy constraints, direct application of the above methods may significantly degrade their performance, which is verified in our experiments. In order to alleviate the noisy side information, Wu et al. proposed a probabilistic framework for metric learning which is however specially designed for RCA [14]. In contrast to this method, we propose a robust optimization framework that is capable of learning from noisy side information. Another important contribution of this paper is that we formulate the related robust optimization problem as a convex programming problem and solve it using efficient Nesterov's smooth optimization method. We show that the proposed methodology can be adapted to the other metric learning methods such as LMNN, D-ranking VM, and the sparse method proposed in [5].

We finally note that our work is closely related to the work of Robust Support Vector Machine (RSVM) [16] which aims to learn a support vector machine from noisily labeled training data. Our work differs from RSVM in that we consider the worst case analysis in our formulation while a best case analysis is utilized in RSVM. As illustrated in [2], unlike the best case analysis, the worst case analysis usually leads to a generalization error bound, making it theoretically sound. Besides, the worst case analysis often results in a convex-concave optimization problem, which can be solved efficiently, while the best case analysis could lead to a non-convex optimization problem that is usually difficult to solve.

## 3 Learning Distance Metric from Perfect Side Information

We first describe the framework of learning a distance metric from perfect side information, which is cast in the form of triplet constraints in our case. In particular, we represent the training examples as a collection of triplets, i.e., $\mathcal{D} = \{(\mathbf{x}_i, \mathbf{y}_i, \mathbf{z}_i), i = 1, \ldots, N\}$, where $\mathbf{x}_i \in \mathbb{R}^d$, $\mathbf{y}_i \in \mathbb{R}^d$, and $\mathbf{z}_i \in \mathbb{R}^d$. In each triplet $(\mathbf{x}_i, \mathbf{y}_i, \mathbf{z}_i)$, $(\mathbf{x}_i, \mathbf{y}_i)$ forms a must-link (or similar) pair and $(\mathbf{x}_i, \mathbf{z}_i)$ form a cannot-link (or dissimilar) pair. We expect $d_A(\mathbf{x}_i, \mathbf{y}_i)$ to be significantly smaller than $d_A(\mathbf{x}_i, \mathbf{z}_i)$, where $d_A(\mathbf{x}, \mathbf{y}) = (\mathbf{x} - \mathbf{y})^\top A(\mathbf{x} - \mathbf{y})$. Assuming all the training triples in $\mathcal{D}$ are correct, we can cast the distance metric learning into the following optimization problem:

$$\min_{A \succeq 0} \quad \frac{\lambda}{2} \|A\|_F^2 + \sum_{i=1}^{N} \ell(d_A(\mathbf{x}_i, \mathbf{z}_i) - d_A(\mathbf{x}_i, \mathbf{y}_i)) \quad (1)$$

where $\ell(z)$ is a pre-defined loss function. The computational challenge of Eq. 1 arises from handling the linear matrix inequality (LMI) $A \succeq 0$. It can be addressed in two ways. First, we could simplify the constraint $A \succeq 0$ by assuming that

$$A \in \mathcal{S} = \left\{ A = \sum_{k=1}^{m} \lambda_k \xi_k \xi_k^\top : \lambda = (\lambda_1, \ldots, \lambda_m) \in \mathbb{R}_+^m \right\}$$

where $\xi_k \in \mathbb{R}^d, k = 1, \ldots, m$ are base vectors that are derived from the existing data points. In this way, we approximate the original Semi-Definite Programming (SDP) problem into a quadratic optimization problem, and therefore can be solved efficiently. We can further improve this scheme by greedily searching for the base vector $\xi_k$. Let $A$ be the current solution, and we aim to update $A$ by $A + \alpha \xi \xi^\top$ with unknown $\alpha$ and $\xi \in \mathbb{R}^d$. We choose the base by maximizing the gradient of the objective function, i.e.,

$$\min_{|\xi|_2 = 1} \xi^\top Z \xi \,,$$

where $Z$ is defined as

$$Z = \lambda A + \sum_{i=1}^{N} \partial \ell(d_A(\mathbf{x}_i, \mathbf{z}_i) - d_A(\mathbf{x}_i - \mathbf{y}_i)) K_i,$$

with $K_i = (\mathbf{x}_i - \mathbf{z}_i)(\mathbf{x}_i - \mathbf{z}_i)^\top - (\mathbf{x}_i - \mathbf{y}_i)(\mathbf{x}_i - \mathbf{y}_i)^\top$. Hence, $\xi$ is the minimum eigenvector of matrix $Z$.

The second approach is to learn the distance metric $A$ by an online learning algorithm by assuming that the training triples are provided sequentially. The details can be found in [6].

## 4 Learning Distance Metric from Noisy Side Information

We now turn to the case of learning distance metric from noisy side information. Let's assume that we know a priori that at most $1 - \eta \in [0, 1)$ of the triples in $\mathcal{D}$ are uncertain. The value $\eta$ can be estimated with reasonable accuracy by randomly sampling the triples from $\mathcal{D}$ and checking if the labels are correct. We emphasize that although the percentage of noisy or uncertain triples can be reliably estimated, we do not know which subset of the triples in $\mathcal{D}$ are uncertain, making it a challenging learning problem.

### 4.1 Robust Model
In order to address the uncertainty in correct triples, we explore the idea of robust optimization [2]. It searches for a distance metric $A$ that minimizes the loss function for any $\eta$ percent of the triples in $\mathcal{D}$, i.e.,

$$\min_{t, A \succeq 0} t + \frac{\lambda}{2} \|A\|_F^2 \qquad \text{s.t.} \qquad (2)$$

$$t \geq \sum_{i=1}^{N} q_i \ell(d_A(\mathbf{x}_i, \mathbf{z}_i) - d_A(\mathbf{x}_i, \mathbf{y}_i)) \,\forall \mathbf{q} \in \mathcal{Q}(\eta)$$

where set $\mathcal{Q}$ is defined as follows

$$\mathcal{Q}(\eta) = \left\{ \mathbf{q} \in \{0, 1\}^N : \sum_{i=1}^{N} q_i \leq N\eta \right\}.$$

As indicated by Eq. 2, the learned distance metric $A$ will minimize the loss of any possible subset of training triples whose cardinality is no more than $N\eta$. Note that the number of constraints in (2) is exponential in $N\eta$, which makes it computationally infeasible. The following lemma allows us to simplify (2) by replacing the combinatorial set $\mathcal{Q}(\eta)$ with its convex hull $\widehat{\mathcal{Q}}$.

**Lemma 1.** *Eq. 2 is equivalent to the following optimization problem*

$$\min_{t, A \succeq 0} t + \frac{\lambda}{2} \|A\|_F^2 \qquad \text{s.t.} \qquad (3)$$

$$t \geq \sum_{i=1}^{N} q_i \ell(d_A(\mathbf{x}_i, \mathbf{z}_i) - d_A(\mathbf{x}_i, \mathbf{y}_i)) \,\forall \mathbf{q} \in \widehat{\mathcal{Q}}(\eta)$$

*where $\widehat{\mathcal{Q}}(\eta)$ is defined as follows*

$$\widehat{\mathcal{Q}}(\eta) = cl(\mathcal{Q}(\eta)) = \left\{ \mathbf{q} \in [0, 1]^N : \sum_{i=1}^{N} q_i \leq N\eta \right\}$$

Lemma 1 follows directly the fact that Eq. 3 is linear in $q$, and as a result, its optimum should be achieved at the extreme points.

Eq. 3 can be viewed as a semi-infinite programming problem [4], since the number of constraints is infinite. For the rest of the paper, we assume a hinge loss for $\ell(z)$, i.e., $\ell(z) = \max(1 - z, 0) = (1 - z)_+$. We have the following lemma that transforms the semi-infinite programming problem in Eq. 3 into a convex-concave optimization problem.

**Lemma 2.** *Eq. 3 is equivalent to the following optimization problem*

$$\max_{A \succeq 0} \min_{\mathbf{q} \in \widehat{\mathcal{Q}}(\eta)} -\frac{\lambda}{2} \|A\|_F^2 - \sum_{i=1}^{N} q_i (1 - \text{tr}(AK_i))_+ \qquad (4)$$

*where $K_i = (\mathbf{x}_i - \mathbf{z}_i)(\mathbf{x}_i - \mathbf{z}_i)^\top - (\mathbf{x}_i - \mathbf{y}_i)(\mathbf{x}_i - \mathbf{y}_i)^\top$.*

The following lemma allows us to simplify the convex-concave optimization Eq. 4 into a maximization problem.

**Lemma 3.** *The convex-concave optimization problem in (4) is equivalent to the following minimization problem*

$$\min_{\mathbf{q}\in\widehat{\mathcal{Q}}(\eta)} \mathcal{L}(\mathbf{q}) \triangleq -\sum_{i=1}^{N} q_i + \frac{1}{2\lambda}\left|\pi_{S_+}\left(\sum_{i=1}^{N} q_i K_i\right)\right|_F^2, \quad (5)$$

*where $\pi_{S_+}(B)$ projects the matrix $B$ onto the SDP cone.*

Lemma 3 can be verified by simply setting the first order derivative with respect to $A$ to be zero.

We could solve (5) by a sub-gradient approach. The main difficulty in deploying the sub-gradient approach is to compute the sub-gradient of the projection function $\pi_{S_+}(A)$. The following lemma allows us to compute the sub-gradient effectively.

**Lemma 4.** *The gradient of the objective function Eq. 5, denoted by $\partial \mathcal{L}(\mathbf{q})$, is computed as*

$$\partial \mathcal{L}(\mathbf{q}) = -\mathbf{1} + \frac{1}{\lambda}\mathbf{h} \quad (6)$$

*where $\mathbf{h} = (h_1, \ldots, h_N)$ and each $h_i$ is computed as*

$$h_i = \text{tr}\left(K_i \pi_{S_+}\left(\sum_{i=1}^{N} q_i K_i\right)\right) \quad (7)$$

*where $\pi_{S_+}(A)$ projects the matrix $A$ onto the SDP cone.*

Using Eq. 6, we can adopt the Rosen gradient method [10], which projects the sub-gradient to domain $\widehat{\mathcal{Q}}(\eta)$ and solve the optimization problem in Eq. 6 iteratively. In each step, the best step size can be obtained by an efficient line search. However, it is well known that the sub-gradient method is usually slow in convergence, particularly for non-smooth objective function. In the following, we aim to improve the learning efficiency by using Nesterov's smooth optimization method [7].

### 4.2 Smooth Optimization

We intend to solve the optimization problem in Eq. 5 using the smooth optimization method [7]. First, it is easy to verify that the objective function in (5) is Lipschitz continuous with respect to $L_2$ norm of $\mathbf{q}$, with Lipschitz constant $L$ bounded by

$$L \leq \sqrt{\sum_{i=1}^{N}\left(1 + \frac{1}{\lambda}|h_i|\right)^2} \leq \sqrt{\sum_{i=1}^{N}\left[1 + \frac{1}{\lambda}\text{tr}(K_i Z)\right]^2} \quad (8)$$

where $Z = \sum_{i=1}^{N}(\mathbf{x}_i - \mathbf{z}_i)(\mathbf{x}_i - \mathbf{z}_i)^\top$.

We can deploy the smooth optimization approach to optimize Eq. 5, which has convergence rate of $O(1/\sqrt{\varepsilon})$, where $\varepsilon$ is the target error. This is significantly better when compared to the convergence rate of sub-gradient descent (i.e., $O(1/\epsilon^2)$). Algorithm 1 shows the smooth optimization algorithm for Eq. 5. There are two optimization problems when applying Algorithm 1, i.e., the problems in Eq. 10 and 11. Both problems have the following general form

$$\min_{\mathbf{q}} \quad \frac{L}{2}|\mathbf{q} - \mathbf{a}|_2^2 + (\mathbf{q} - \mathbf{b})^\top \mathbf{s} \quad (16)$$
$$\text{s. t.} \quad \mathbf{q} \in [0,1]^N, \mathbf{q}^\top \mathbf{1} \leq N\eta$$

where $\mathbf{a}$, $\mathbf{b}$, and $\mathbf{s}$ are given vectors. Below, we discuss how to efficiently optimize the problem in Eq. 16. First, it is straightforward to verify that Eq. 16 can be cast into the following convex-concave optimization problem

$$\max_{\lambda \geq 0} \min_{\mathbf{q} \in [0,1]^N} \frac{L}{2}|\mathbf{q} - \mathbf{a}|_2^2 + (\mathbf{q} - \mathbf{b})^\top \mathbf{s} + \lambda(\mathbf{q}^\top \mathbf{1} - N\eta).$$

By solving the minimization problem, we have $\mathbf{q}$ expressed as

$$q_i = \pi_{[0,1]}(a_i - s_i/L - \lambda)$$

According to KKT condition, we have $\lambda(\mathbf{q}^\top \mathbf{1} - N\eta) = 0$. Hence, to decide the value for $\lambda$, we have the following procedure

- We first try $\lambda = 0$. If $\sum_{i=1}^{N} q_i \leq N\eta$ when $\lambda = 0$, that is the optimal solution.
- If not, we search $\lambda > 0$ that $\mathbf{q}^\top \mathbf{1} = N\eta$. Since $q_i$ is monotonically decreasing in $\lambda$, a bi-section search can be used to identify $\lambda$ that satisfies $\mathbf{q}^\top \mathbf{1} = N\eta$.

The theorem below shows the convergence rate of the smooth optimization algorithm described in Algorithm 1

**Theorem 1.** *After running the algorithm specified in Algorithm 1 for $T$ iterations, we have the following inequality holds*

$$|\mathcal{L}(\mathbf{q}_T) - \mathcal{L}(\mathbf{q}^*)| \leq \frac{2LN\eta}{(T+1)(T+2)}$$

*where $\mathbf{q}^* = \arg\max_{\mathbf{q}\in\widehat{\mathcal{Q}}(\eta)} \mathcal{L}(\mathbf{q})$.*

The proof follows directly Theorem 2 in [7]. The theorem below allows us to bound the maximum number of iteration for the algorithm in Algorithm 1 to quit.

**Algorithm 1** Smooth Optimization Algorithm for Solving (5)

1: **Input**
   - $\varepsilon$: predefined error for optimization
2: **Initialization**
   - Compute the Lipschitz constant $L$ using Eq. 8
   - Initialize the solution $\mathbf{z}_0 = \mathbf{0} \in \mathbb{R}^N$
   - $t = 0$
3: **repeat**
4:     Compute the solution for $\mathbf{q}_t$
   - Computes

$$\hat{A}_t = \frac{1}{\lambda} \pi_{S_+}\left(\sum_{i=1}^{N} z_{t,i} K_i\right), \ \nabla \mathcal{L}(\mathbf{z}_t) = -\mathbf{1} + \mathbf{h}/\lambda, \ \mathcal{L}(\mathbf{z}_t) = -\mathbf{z}_t^\top \mathbf{1} + \frac{\lambda}{2}|\hat{A}_t|_F^2 \tag{9}$$

   where $\pi_{S_+}(M)$ projects matrix $M$ into the SDP cone and $\mathbf{h} = (h_1, \ldots, h_N)$ with $h_i = \text{tr}(K_i \hat{A}_t)$.
   - Solve the following optimization problem

$$\hat{\mathbf{q}} = \arg\min_{\mathbf{q} \in \hat{\mathcal{Q}}(\eta)} \left\{ \frac{L}{2}|\mathbf{q} - \mathbf{z}_t|_2^2 + (\mathbf{q} - \mathbf{z}_t)^\top \nabla \mathcal{L}(\mathbf{q}) \right\} \tag{10}$$

   - $\mathbf{q}_t = \arg\min_{\mathbf{q}'}\{\mathcal{L}(\mathbf{q}') : \mathbf{q}' \in \{\mathbf{q}_{t-1}, \hat{\mathbf{q}}, \mathbf{z}_t\}\}$

6:     Compute $\mathbf{z}_{t+1}$
   - Solve the following optimization problem

$$\mathbf{w}_t = \arg\min_{\mathbf{q} \in \hat{\mathcal{Q}}(\eta)} \left\{ \frac{L}{2}|\mathbf{q}|_2^2 + \sum_{j=0}^{t} \frac{j+1}{2}(\mathbf{q} - \mathbf{z}_j)^\top \nabla \mathcal{L}(\mathbf{z}_j) \right\} \tag{11}$$

   - Set

$$\mathbf{z}_{t+1} = \frac{2}{t+3}\mathbf{w}_t + \frac{t+1}{t+3}\mathbf{q}_t \tag{12}$$

7:     Compute the duality gap $\Delta$
   - Set

$$A_t = \frac{4}{(t+1)(t+2)} \sum_{j=0}^{t} \frac{j+1}{2} \hat{A}_j \tag{13}$$

   - Compute duality gap $\Delta_i$ as

$$\Delta_t = \mathcal{L}(\mathbf{q}_t) - \mathcal{H}(A_t) \tag{14}$$

   where $\mathcal{H}(A_t)$ is computed as

$$\mathcal{H}(A_t) = -\frac{\lambda}{2}\|A_t\|_F^2 - \max_{\mathbf{q} \in \hat{\mathcal{Q}}(\eta)} \sum_{i=1}^{N} q_i(1 - \text{tr}(A_t K_i)) \tag{15}$$

8: **until** $\Delta_t \leq \varepsilon$

**Theorem 2.** *The number of iterations for running the algorithm in Algorithm 1 before it quits, denoted by $T$, is bounded as follows*

$$T < \sqrt{\frac{2LN\eta}{\varepsilon}} - 1$$

*Proof.* First, we have

$$\sum_{t=0}^{T} \frac{t+1}{2} [\mathcal{L}(\mathbf{z}_t) + (\mathbf{q} - \mathbf{z}_t)^\top \nabla \mathcal{L}(\mathbf{z}_t)]$$

$$\leq \sum_{t=0}^{T} \frac{t+1}{2} \left\{ -\mathbf{z}_t^\top \mathbf{1} - \frac{\lambda}{2} |\hat{A}_t|_F^2 + \sum_{i=1}^{N} z_{t,i} \text{tr}(\hat{A}_t K_i) - (\mathbf{q} - \mathbf{z}_t)^\top (\mathbf{1} - \mathbf{h}/\lambda) \right\}$$

$$= \sum_{t=0}^{T} \frac{t+1}{2} \left[ -\mathbf{q}^\top \mathbf{1} - \frac{\lambda}{2} \|\hat{A}_t\|_F^2 + \sum_{i=1}^{n} q_i \text{tr}(\hat{A}_t K_i) \right]$$

$$\leq Q(T) \left( -\mathbf{q}^\top \mathbf{1} - \frac{\lambda}{2} |A_T|_F^2 + \sum_{i=1}^{N} q_i \text{tr}(\hat{A}_t K_i) \right)$$

In the above, $Q(T) = \frac{(T+1)(T+2)}{4}$. The last step uses the concaveness of $\mathcal{H}(A)$. Using Theorem 2 of the work by Nesterov [7], we have

$$\mathcal{L}(\mathbf{q}_T) \leq \frac{2LN\eta}{(T+1)(T+2)} +$$

$$\min_{\mathbf{q} \in \hat{\mathcal{Q}}(\eta)} \left\{ -\mathbf{q}^\top \mathbf{1} - \frac{\lambda}{2} \|A_T\|_F^2 + \sum_{i=1}^{N} q_i \text{tr}(\hat{A}_t K_i) \right\}$$

$$\leq \frac{2L_g N\eta}{(T+1)(T+2)} + \mathcal{H}(A_T)$$

Since $\Delta = \mathcal{H}(A_T) - \mathcal{L}(\mathbf{q}_T)$, we have the result in the theorem. □

### 4.3 Extension to Other Metric Learning Methods

As shown by Huang et al. [5], many distance metric learning optimization problems can be seen as an optimization problem with different regularization terms. For instance, LMNN can be formulated as a similar problem regularized by a trace form, i.e.,

$$\min_{A \succeq 0} \sum_{i=1}^{N} \ell(d_A(\mathbf{x}_i, \mathbf{z}_i) - d_A(\mathbf{x}_i, \mathbf{y}_i)) + \frac{\lambda}{2} \text{tr}(WA),$$

where $W = \frac{1}{N} \sum_{i=1}^{N} (\mathbf{x}_i - \mathbf{y}_i)(\mathbf{x}_i - \mathbf{y}_i)^\top$. Using the similar techniques described in this work, we can extend LMNN to deal with noisy class assignments. Similarly, we can extend Generalized Sparse Metric Learning [5] and D-ranking Vector Machine [8] to their robust versions.

Table 1: Descriptions of UCI data sets

| Data Set. | # Samples | #Dimension | # Class |
|---|---|---|---|
| Sonar | 208 | 60 | 2 |
| Iris | 150 | 4 | 3 |
| Ionosphere | 351 | 34 | 2 |
| Heart | 270 | 13 | 2 |
| Wine | 178 | 13 | 3 |

## 5 Experiments

**Experimental setup** The five UCI data sets are used in the experiments.[1] Table 1 summarizes the statistics of these data sets. We compare the proposed robust metric learning method, denoted as **RML**, with three state-of-the-art algorithms for metric learning, including **LMNN** [12], **Xing**'s method. [15], and the common Euclidean distance metric (**EUCL**). Depending on different optimization methods, we name the RML method as **RML**$_{gd}$ when using sub-gradient optimization and **RML**$_{nes}$ when using Nesterov optimization.

We first apply these metric learning methods to learn a distance metric from the training data, and then deploy the $k$-Nearest Neighbor ($k$NN) method to evaluate the quality of the learned metric. We follow the same setting as in [12] and use the category information to generate the side information. More specifically, in the training data, given $\mathbf{y}$ is $\mathbf{x}$'s $k$ nearest neighbor, we have $\mathbf{x}$ and $\mathbf{y}$ form a must-link pair if they share the same class label and a cannot-link if they belong to different classes. We split each data set randomly into a training set and a test set with 85% of data used for training. We perform training on the triplet set obtained from the training set and evaluate the classification performance of $k$NN in the test set. In order to generate the noisy side information, we randomly switch the correct triplet $(\mathbf{x}_i, \mathbf{y}_i, \mathbf{z}_i)$ to $(\mathbf{x}_i, \mathbf{z}_i, \mathbf{y}_i)$ with the probability $1 - \eta$.

All the metric learning methods involve a trade-off parameter $\lambda$. It is tuned from the range $\{2^{-3}, 2^{-2}, 2^{-1}, 2^0, 2^1, 2^2, 2^3, 2^4, 2^5, 2^6, 100\}$. The parameter $k$ is empirically set to 5 according to our experiments. We repeat the experiments 10 times and then report the results averaged over 10 runs.

**Experimental Results** Table 2 reports the classification errors of $k$NN using the distance metric learned from noisy constraints with $\eta = 0.8$ (i.e., 20% of constraints are uncertain). We observe that the proposed robust framework consis-tently performs better than the baseline algorithms. We also observe that $\text{RML}_{nes}$ and $\text{RML}_{gd}$, the two implementations of the proposed framework using different optimization strategies, yield almost identical performance. For certain data sets, we observe that LMNN and Xing can be affected significantly by the noisy constraints, and learn a distance metric that is significantly worse than the Euclidean distance metric (i.e., EUCL). The degradation is significant for Xing's method in Heart, and for LMNN in Sonar. These results clearly indicate the negative effect caused by noisy constraints. In contrast, the proposed method always outperforms the baseline Euclidean distance metric given the noisy constraints, indicating that our method is able to deal with the uncertain issue appropriately and consequently lift up the performance.

---
[1] http://www.ics.uci.edu/∼mlearn/MLRepository.html.

Table 2: Test error for the proposed algorithm and three baseline algorithms with $\eta = 0.8$.

| Data Set | $\text{RML}_{nes}$ | $\text{RML}_{gd}$ | LMNN | Xing | EUCL |
|---|---|---|---|---|---|
| Sonar (%) | **17.42 ± 5.73** | 17.96 ± 5.25 | 36.45 ± 7.91 | 17.42 ± 6.31 | 19.68 ± 6.17 |
| Iris (%) | **1.82 ± 3.18** | 1.97 ± 3.53 | 2.27 ± 3.21 | 2.73 ± 2.36 | 2.73 ± 2.58 |
| Ionosphere (%) | **11.13 ± 4.12** | **11.13 ± 4.12** | 11.89 ± 4.27 | 13.77 ± 3.78 | 14.15 ± 4.81 |
| Heart (%) | 16.00 ± 7.38 | **15.93 ± 6.32** | 17.50 ± 5.00 | 19.75 ± 7.12 | 17.55 ± 6.39 |
| Wine (%) | **1.48 ± 2.59** | 1.81 ± 2.52 | 6.67 ± 5.74 | 2.96 ± 1.56 | 4.07 ± 4.43 |

To show the efficiency of Nesterov's smooth optimization method, in Fig. 1 (the first 5 graphs), we plot the convergence curves for both Nesterov's optimization method and the sub-gradient method on all data sets. Evidently, Nesterov's method converges significantly faster than the sub-gradient method. This once again verifies the advantage of the smooth optimization method. All the algorithms are implemented and run using Matlab. All the experiments are conducted on Intel Core (TM)2 2.66Ghz CPU machine with 4G RAM and Windows XP OS.

In order to see how $\eta$ affects the performance of different metric learning methods, we plot their test errors against different $\eta$ on the Wine data set in the last graph of Fig. 1. For convenience, the parameter $\lambda$ is set to $2^6$ for all the metric learning methods, while $k$ is still set to 5. It is interesting to note that, $\text{RML}_{nes}$ is quite robust against the noise if noise does not dominate side information. With heavy noise in side information, it may fail to find a better metric than EUCL. This is understandable, since too heavy noise may "bury" the latent best metric and consequently makes restoration less possible. In addition, LMNN appears quite sensitive to noisy side information in this data set; Xing also demonstrates robust property, but performs worse than $\text{RML}_{nes}$ when $\eta$ is reasonably large.

## 6 Conclusion

In this paper, we proposed a robust distance metric learning framework, which is formulated in a worst-case scenario. In contrast to most existing studies that usually assume perfect side information, we developed a novel method that learns from noisy or uncertain side information. We formulated the learning task initially as a combinatorial optimization problem, and then showed that it can be transformed into a convex programming problem. Finally, we exploited the efficient Nesterov smooth optimization method to solve the problem. One appealing feature of the proposed framework is that the robust framework can be adapted to other metric learning methods in order to handle noisy constraints. Experiments on several UCI data sets showed that the proposed novel framework is highly effective for dealing with noisy side information.

## Acknowledgements

The work of Rong Jin was supported by National Science Foundation (IIS-0643494) and National Health Institute (1R01GM079688-01). The work of Kaizhu Huang and Cheng-Lin Liu were supported by the National Natural Science Foundation of China (NSFC) under grant no.60825301.

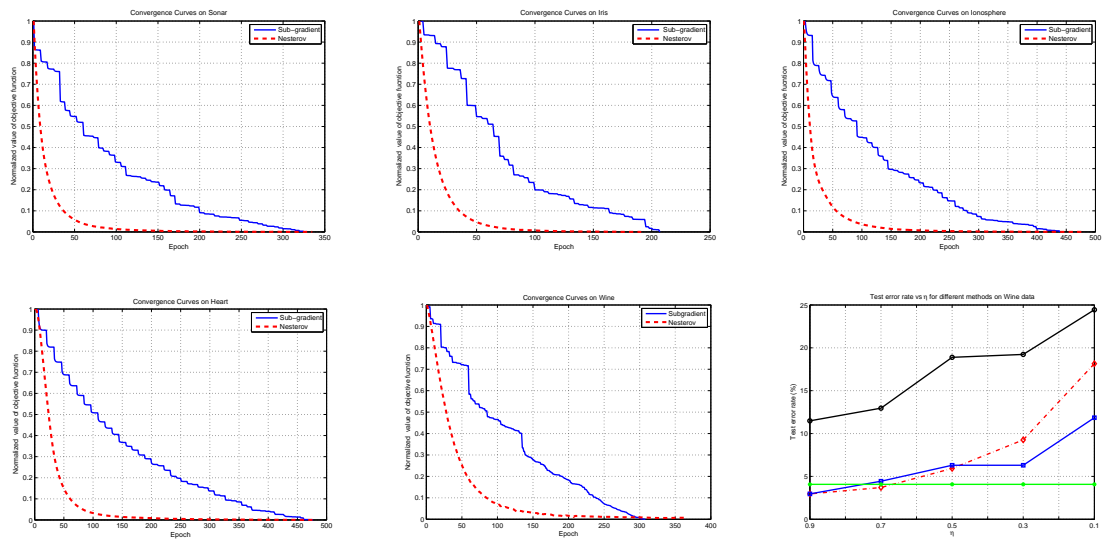

Figure 1: The first 5 graphs: convergence curves for the sub-gradient method and Nesterov's method on five data sets. The right-bottom graph: test error rate vs $\eta$ for different methods on Wine Data.